\documentclass[10pt,twocolumn,letterpaper]{article}

\usepackage[pagenumbers]{cvpr}

\usepackage{graphicx}
\usepackage{amsmath}
\usepackage{amssymb}
\usepackage{booktabs}

\usepackage[utf8]{inputenc}
\usepackage[T1]{fontenc}
\usepackage{booktabs}
\usepackage{amsfonts}
\usepackage{nicefrac}
\usepackage{microtype}
\usepackage{xcolor}
\usepackage{makecell}
\usepackage{graphicx}
\usepackage{multirow}
\usepackage{amsmath}
\usepackage{subcaption}
\usepackage[scaled=0.85]{beramono}
\usepackage{tablefootnote}
\usepackage[pagebackref,breaklinks,colorlinks]{hyperref}

\usepackage[capitalize]{cleveref}
\crefname{section}{Sec.}{Secs.}
\Crefname{section}{Section}{Sections}
\Crefname{table}{Table}{Tables}
\crefname{table}{Tab.}{Tabs.}

\begin{document}

\title{Label-Free Synthetic Pretraining of Object Detectors}

\author{
Hei Law \\
Princeton University
\and
Jia Deng \\
Princeton University
}
\maketitle

\begin{abstract}
    We propose a new approach, Synthetic Optimized Layout with Instance Detection (SOLID), to pretrain object detectors with synthetic images. Our ``SOLID'' approach consists of two main components: (1) generating synthetic images using a collection of unlabelled 3D models with optimized scene arrangement; (2) pretraining an object detector on ``instance detection'' task---given a query image depicting an object, detecting all instances of the exact same object in a target image. Our approach does not need any semantic labels for pretraining and allows the use of arbitrary, diverse 3D models.  Experiments on COCO show that with optimized data generation and a proper pretraining task, synthetic data can be highly effective data for pretraining object detectors. In particular, pretraining on rendered images achieves performance competitive with pretraining on real images while using significantly less computing resources. Code is available at \url{https://github.com/princeton-vl/SOLID}.
\end{abstract}

\section{Introduction}
Object detection is a key computer vision task. Currently, state-of-the-art systems are trained on a large number of manually annotated images. However,  manual annotations are costly to acquire; as a result, such reliance can potentially become a bottleneck for further improvement. An important question is whether additional performance gains can be achieved by using alternative sources of data without manual labels.\looseness=-1

To reduce the dependency on manual  labels, recent work~\cite{DBLP:conf/cvpr/He0WXG20,DBLP:journals/corr/abs-2003-04297,DBLP:conf/icml/ChenK0H20,DBLP:conf/iccv/HenaffKAOVC21,DBLP:conf/cvpr/RohSKK21,DBLP:conf/iccv/XiaoR0KD21,DBLP:journals/corr/abs-2106-04550,DBLP:conf/nips/CaronMMGBJ20,DBLP:conf/cvpr/YeZYC19,DBLP:journals/corr/abs-2111-06377,DBLP:conf/cvpr/PathakKDDE16,DBLP:conf/icml/VincentLBM08,DBLP:journals/jmlr/VincentLLBM10,DBLP:journals/corr/abs-2106-08254,DBLP:conf/icml/ChenRC0JLS20,DBLP:conf/cvpr/MisraM20,DBLP:conf/cvpr/KolesnikovZB19,DBLP:conf/iccv/DoerschZ17,DBLP:conf/nips/0001X0L0020,DBLP:conf/cvpr/ChenH21,DBLP:conf/iclr/AsanoRV20a,DBLP:conf/eccv/CaronBJD18,DBLP:conf/icml/HuangDGZ19} has explored self-supervised pretraining, which leverages the massive amounts of unlabeled images online. Common approaches of self-supervision include contrastive learning~\cite{DBLP:conf/cvpr/He0WXG20,DBLP:journals/corr/abs-2003-04297,DBLP:conf/icml/ChenK0H20,DBLP:conf/iccv/HenaffKAOVC21,DBLP:conf/cvpr/RohSKK21,DBLP:conf/iccv/XiaoR0KD21,DBLP:journals/corr/abs-2106-04550,DBLP:conf/nips/CaronMMGBJ20,DBLP:conf/cvpr/YeZYC19}, where a network learns features invariant to known 2D image augmentations, and reconstructive learning~\cite{DBLP:journals/corr/abs-2111-06377,DBLP:conf/cvpr/PathakKDDE16,DBLP:conf/icml/VincentLBM08,DBLP:journals/jmlr/VincentLLBM10,DBLP:journals/corr/abs-2106-08254,DBLP:conf/icml/ChenRC0JLS20}, where a network learns to predict missing/masked parts of data using the rest.

Despite many promising results, self-supervised  pretraining still faces significant technical challenges. Contrastive approaches heavily depend on effective image   augmentations, which can be difficult to design beyond a few simple 2D transforms. Reconstructive approaches can be highly sensitive to the relevance of the reconstruction task to downstream applications. For example, reconstructing raw pixel intensities may not be as useful for object detection where invariance to intensity changes is important.\looseness=-1

On the other hand, synthetic data has been widely used to supplement real images for many computer vision tasks, with notable successes in 3D vision~\cite{DBLP:conf/cvpr/MayerIHFCDB16,DBLP:conf/eccv/ButlerWSB12,DBLP:conf/cvpr/TremblayTB18,DBLP:conf/iros/WangZWHQWHKS20,DBLP:conf/3dim/LipsonTD21,DBLP:conf/eccv/ZhangQYPWT20,DBLP:conf/cvpr/ZhangPYT19,DBLP:conf/nips/TeedD21,DBLP:conf/eccv/LabbeCAS20}. However, using synthetic data has yet to become common practice for object detection, except in specialized domains such as autonomous driving~\cite{DBLP:conf/cvpr/TremblayPABJATC18,DBLP:conf/icra/Johnson-Roberson17,DBLP:conf/bmvc/AlhaijaMMGR17,DBLP:conf/iccv/PrakashDLCSBL21}. One possible reason for this discrepancy is that for synthetic data to work well, they should closely approximate the real data (i.e.\@ small real-sim domain gap), but for object detection it is difficult to generate synthetic data that matches real data in terms of the diversity of objects and scenes. For example, it would be extremely challenging to create a synthetic dataset to cover all 80 object categories in the COCO benchmark~\cite{DBLP:conf/eccv/LinMBHPRDZ14}, including realistic compositions of scenes and varied instances of each object category (e.g.\@ varied dogs and cats). Thus it is perhaps expected that outside specialized domain, synthetic data would have a limited role to play for  object detection. 

In this work, we present a new result that defies conventional wisdom: synthetic data can be surprisingly effective for object detection, even with limited diversity and realism. In particular, we can generate effective synthetic data using only a collection of 3D models \emph{without any category labels}; pretraining on such synthetic data achieves results competitive to self-supervised pretraining on real images.\looseness=-1

\begin{figure*}[t]
    \centering
    \resizebox{\textwidth}{!}{\includegraphics{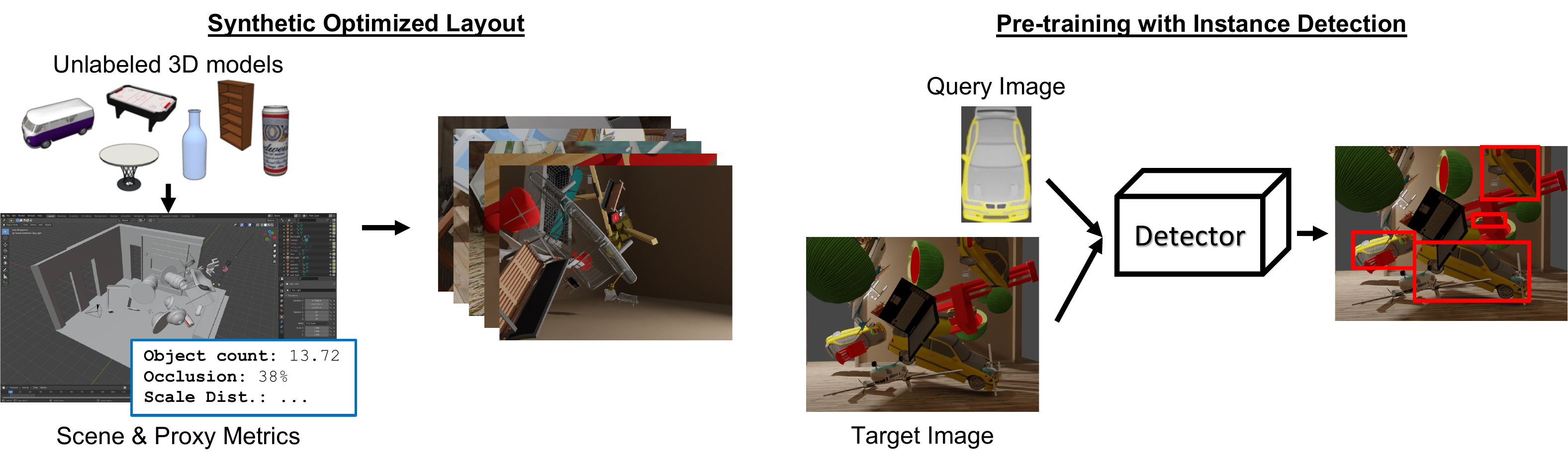}}
    \caption{Overview of our approach. We generate synthetic images using unlabeled 3D models with optimized scene layout. We then pretrain an object detector to perform instance detection: given a query object, detect all instances in a target image. }
    \label{fig:overview}
    \vspace{-3mm}
\end{figure*}

We achieve this result through ``Synthetic Optimized Layout with Instance Detection (SOLID)'', a new approach we introduce for pretraining object detectors. Our approach, as illustrated in Fig.~\ref{fig:overview}, consists of two main ingredients: (1) generating synthetic images using a collection of unlabeled 3D models, with optimized scene arrangement; (2) pretraining an object detector on the ``instance detection'' task~\cite{DBLP:conf/wacv/MercierGGL21,DBLP:journals/corr/abs-1803-04610}---given a query image depicting an object, detecting all instances of the exact same object in a target image.

It is worth noting that our approach only uses \emph{unlabeled} 3D models. That is, no semantic labels are necessary; in fact, the 3D models can be arbitrary shapes that do not fit into any known object category. Doing away with semantic labels gets around the difficulty in generating 3D shapes that conform to semantic categories (e.g.\@ generating varied and realistic 3D shapes of cats is an unsolved problem), allowing the use of arbitrary 3D shapes which can  significantly expand the diversity of synthetic data. Coupled with this label-free synthetic data is the instance detection pretraining task, which, unsurprisingly, also requires no semantic labels. The task is posed purely geometrically---finding in a target image all instances of the 3D shape depicted in a query image. 

Our instance detection task differs from the standard object detection in that the input consists of two images, a query image that specifies the object to detect, and a target image in which to detect the instances of the object. The object specified by the query image may be completely novel and may have never been seen during training.  In other words, unlike standard object detection, there is no fixed, predetermined list of objects or object classes to detect. During test time, the network needs to detect all the instances given just a single example of a new object. 

Our ``SOLID'' approach is motivated by the hypothesis that a significant source of difficulty of object detection is geometrical invariances, particularly invariances to occlusion, illumination, and viewpoint. Such invariances are likely learnable independent of category labels---an infant may not know that the object she likes to play with is called a ``toy car'', but she can almost certainly find it in a photo effortlessly. This hypothesis naturally leads to the instance detection task, which focuses the pretraining on learning geometric invariances.
 
Given a collection of unlabeled 3D models, there are infinitely many possible synthetic images one can generate, but only a finite subset of them can be used during pretraining and not all subsets are equally useful. We thus optimize the parameters and design choices of the rendering pipeline to maximize the effectiveness of the synthetic data. This primarily involves the spatial layout of the objects in front of the camera. A naive approach would be to try each possible layout, render a dataset, pretrain a model, fine-tune on labeled data, and evaluate performance on a validation set of real images. But this would be too expensive to be feasible. Instead, we optimize the layout against a set of proxy metrics of scene complexity including object count, amount of occlusion, and scale distribution, which we find to be good indicators of the validation performance on real images. \looseness=-1

The main novelty of work is a new pretraining method of object detection that integrates two existing ideas: synthetic data and instance detection. Neither synthetic data nor instance detection is novel on its own, but to our knowledge no prior work has combined the two for pretraining object detection.
Our approach has unique advantages over existing alternatives. Compared to existing methods using synthetic data, our approach does not require semantic labels, which means that arbitrary 3D shapes can be used. Compared to existing methods using real images, our approach allows 3D data augmentations that are difficult to achieve with real images. \looseness=-1

We evaluate our approach on the standard COCO~\cite{DBLP:conf/eccv/LinMBHPRDZ14} dataset. We generate synthetic images using 3D models from ShapeNet~\cite{DBLP:journals/corr/ChangFGHHLSSSSX15} and SceneNet~\cite{DBLP:journals/corr/HandaPBSC15a}, and pretrain standard object detectors including Faster-RCNN~\cite{DBLP:journals/pami/RenHG017} and Mask-RCNN~\cite{DBLP:conf/iccv/HeGD19}. Experiments on COCO show that pretraining on rendered images achieves performance competitive with pretraining on real images, including MoCov3~\cite{DBLP:conf/iccv/ChenXH21}, DetCon~\cite{DBLP:conf/iccv/HenaffKAOVC21} and SCRL~\cite{DBLP:conf/cvpr/RohSKK21}, while using significantly less computing resources. Our results demonstrate that with our novel combination of optimized data generation and a proper pretraining task, synthetic data can be highly effective data for pretraining object detectors. \looseness=-1

\section{Related Work}
\noindent\textbf{Supervised and Unsupervised pretraining for Object Detection}
Pretraining visual models including object detectors has a long history. Early work~\cite{DBLP:journals/neco/HintonOT06,DBLP:conf/nips/RanzatoPCL06,bengio2007scaling} shows that pretraining each layer with an unsupervised learning algorithm before fine-tuning the network improves the network's performance significantly. A similar approach is also used in a pedestrian detector~\cite{DBLP:conf/cvpr/SermanetKCL13}. Later, R-CNN~\cite{DBLP:conf/cvpr/GirshickDDM14} shows that supervised pretraining on the ImageNet classification dataset~\cite{DBLP:conf/cvpr/DengDSLL009} significantly improves detection performance. Since then, pretraining on the image classification task has become a common practice in state-of-the-art object detectors~\cite{DBLP:journals/corr/abs-2103-07461,DBLP:conf/cvpr/SunZJKXZTLYW021,DBLP:conf/cvpr/TanPL20}.

Recent approaches~\cite{DBLP:conf/cvpr/He0WXG20,DBLP:journals/corr/abs-2003-04297,DBLP:conf/iccv/ChenXH21,DBLP:conf/icml/ChenK0H20,DBLP:conf/nips/ChenKSNH20,DBLP:conf/nips/CaronMMGBJ20} based on contrastive learning~\cite{DBLP:conf/cvpr/HadsellCL06} has achieved competitive performance against  supervised pretraining on various downstream tasks. In contrastive learning, a network learns to predict similar embeddings for augmented views of the same image. But these approaches rely on image-level features to predict the embeddings, which is not ideal for object detection. Newer approaches have proposed new pretraining tasks that focus on the object-level features. SCRL~\cite{DBLP:conf/cvpr/RohSKK21}, $\text{ReSim-FPN}^\text{T}$~\cite{DBLP:conf/iccv/XiaoR0KD21} and DetCo~\cite{DBLP:conf/iccv/XieDWZXSL021} train a network to predict similar embedding vectors between patches of the two augmented views. Instead of patches, DetCon~\cite{DBLP:conf/iccv/HenaffKAOVC21} uses regions that are segmented by a graph-based algorithm. InsLoc~\cite{DBLP:conf/cvpr/YangWZL21} randomly crops two overlapping patches from an image, paste them onto different images and train a network to predict similar embedding vectors between the patches.

UP-DETR~\cite{DBLP:conf/cvpr/DaiCLC21} proposes a pretraining task where random patches are extracted from an image and a network is trained to localize them in the same image. DETReg~\cite{DBLP:journals/corr/abs-2106-04550} proposes to train an object detector  to predict bounding boxes generated from the selective search algorithm~\cite{DBLP:journals/ijcv/UijlingsSGS13} and to mimic the output of a network trained with contrastive learning.

Our approach is similar to these existing approaches in that our pretraining task can be thought of  as a form of contrastive learning---predicting embeddings to distinguish image regions of the same object from other regions. But our approach differs  in that we use synthetic images instead of real images.  Using synthetic images allows us to easily generate augmented versions of the same object with occlusion and viewpoint change. Such augmentations would be very difficult to generate for natural images.\looseness=-1

\noindent \textbf{Instance Detection}
The instance detection task has also been studied by prior work. Ammirato et al.~\cite{DBLP:journals/corr/abs-1803-04610} and Mercier et al.~\cite{DBLP:conf/wacv/MercierGGL21} suggest this task is useful for robotics and augmented reality applications which often require recognizing a very specific instance, and propose different approaches to the task. Compared to these works,  the novelty of our work is in integrating synthetic data and the instance detection task to pretrain object detectors.

\noindent
\textbf{Synthetic Training of Object Detectors}
Prior work has also studied how to use synthetic data to train an object detector. Peng et al.~\cite{DBLP:conf/iccv/PengSAS15}, Alhaija et al.~\cite{DBLP:conf/bmvc/AlhaijaMMGR17}, Hinterstoisser et al.~\cite{DBLP:conf/eccv/HinterstoisserL18} and Tremblay et al.~\cite{DBLP:conf/cvpr/TremblayPABJATC18} generate the synthetic data by rendering 3D models over a real world background scene. The synthetic data are then used to fine-tune networks pretrained on ImageNet. 

Other work has obtained training data from computer games for object detection and semantic segmentation. Bochinski et al.~\cite{DBLP:conf/avss/BochinskiES16} use Garry's Mod to train a detector for cars, persons and animals. Richter et al.~\cite{DBLP:conf/eccv/RichterVRK16} obtain data for semantic segmentation in Grand Theft Auto. Shafaei et al.~\cite{DBLP:conf/bmvc/ShafaeiLS16} show that a network pretrained on synthetic data collected from a game outperforms network pretrained on real world data after fine-tuning. Data extracted from the games may not come with foreground or background labels. So extra steps such as background subtraction~\cite{DBLP:conf/avss/BochinskiES16} or manual labelling~\cite{DBLP:conf/eccv/RichterVRK16} are needed to identify foreground objects or pixels.

Our method also uses synthetic data, but unlike these existing approaches,  we introduce a new pretraining task. All of these prior works perform pretraining in the form of standard object detection---the input consists of a single image and the task is detect objects in a fixed, pre-defined; as a result, the network needs to memorize through training what each object in the list looks like. 
In contrast, our instance detection pretraining task has two input images (a query image and a target image), and the network only needs to learn geometric invariances as opposed to the object-specific visual features which are less transferrable to new domains.

\section{Approach}
Our ``SOLID'' approach involves two main ingredients: generating synthetic images and pretraining a network on instance detection--given a query image of an unknown  object, detecting all instances of the same object in a target image. 
After pretraining, the pretrained object detector is fine-tuned on a downstream dataset. Fig.~\ref{fig:overview} gives an overview of SOLID.

\begin{figure*}
    \centering
    \resizebox{0.8\textwidth}{!}{\includegraphics{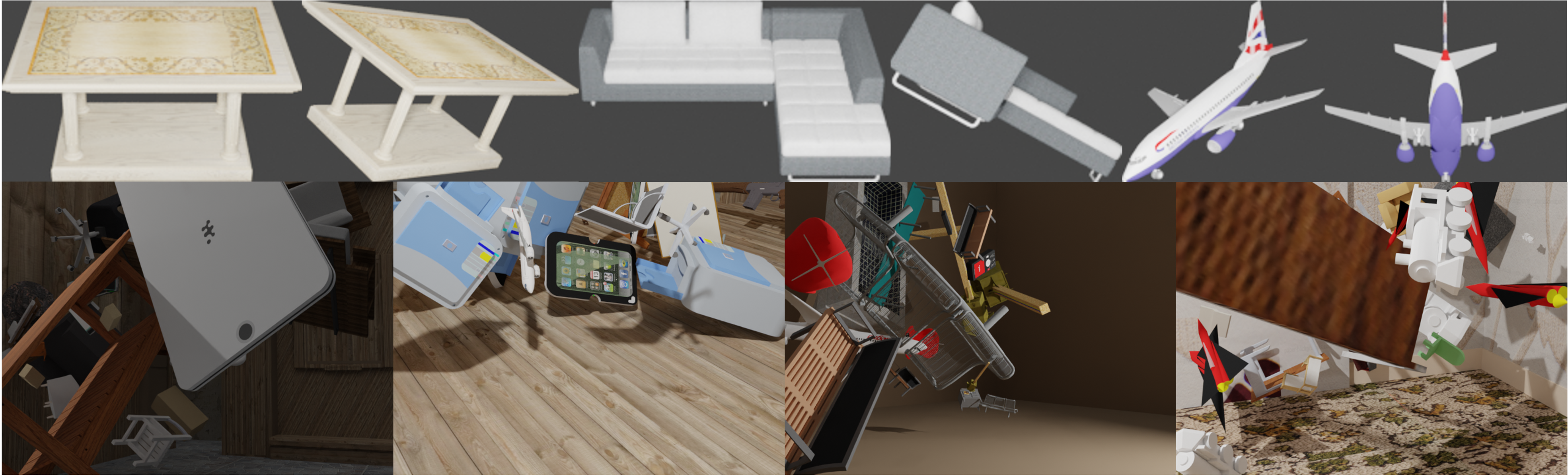}}
    \caption{The first row shows query images and the second row shows target images.}
    \label{fig:images}
\end{figure*}

\subsection{Generating Synthetic Images}
We generate our synthetic images by placing 3D models from ShapeNet~\cite{DBLP:journals/corr/ChangFGHHLSSSSX15} into backgrounds  from SceneNet~\cite{DBLP:journals/corr/HandaPBSC15a}. There is a large space of possible layouts of the objects relative to the camera, but some of them are likely to be more useful than others as synthetic data. We can find out the usefulness of a subset of layouts by using them to pretrain a detector and evaluate the validation performance on real images, but this would be prohibitively expensive. Instead, we propose a set of proxy metrics of scene complexity to guide data generation. These proxy metrics can be evaluated quickly without going through actual pretraining and  capture the known failure cases of object detection (crowded scenes, occlusion, small objects, viewpoint change, etc.). Empirically we find these metrics to correlate well with the final validation performance. 

\noindent \textbf{Proxy Metrics}~~We use the following proxy metrics of scene complexity: 
\begin{itemize}
    \item \textbf{Average occlusion:} For each object, in addition to the usual segmentation mask $m_p$ used in the instance segmentation task, we render a segmentation mask $m_{f}$ by hiding other objects in the scene. The occlusion of an object is defined as $(n_{f} - n_{p}) / n_{f}$ where $n_{p}$ is the number of pixels in $m_{p}$ and $n_{f}$ is the number of pixels in $m_{f}$.
    \item \textbf{Scale distributions:} Following the definition in COCO, we divide the objects into small, medium and large objects by the number of pixels in their segmentation masks. We then calculate the percentage of each scale.
    \item \textbf{Object count:} This is the number of visible objects in the rendered image. Some objects may not be visible in the rendered image because they may be completely occluded.
\end{itemize}

\noindent \textbf{Rendering Pipeline}~~The scenes from SceneNet come with 3D models. We remove all 3D models in the scenes except the ceiling, floor and wall before they are used as backgrounds.

After initializing the background, we randomly place a camera away from the center of the background and point it toward the central area of the background to allow enough space for placing the 3D models. Because there may be walls in the background, we perform ray casting to find out the nearest obstacle to the camera. If it is too close, we sample another location until there is enough space to place the objects. Only the locations within the viewing frustum  is considered in the subsequent steps, which improves efficiency and ensures that at least part of the objects can be visible in the final images. We randomly select, scale, and rotate a 3D model and place it at a location where its 3D bounding box does not intersect with 3D bounding boxes from other models.

In our rendering pipeline, the spatial layout of objects is influenced by several parameters and design choices, which we optimize against our proxy metrics. Instead of randomly placing a 3D model, we place it in front of or behind other 3D models with respect to the camera to increase the amount of occlusion. To increase the object count, we pack more objects into a scene by allowing them to be floating in the air. We optimize the scales and the rotations of the 3D models to have different scale distributions and include more viewpoints for each object.

After placing all the objects, we apply three point lighting to a random object in the scene to illuminate the scene before we render the image. In all of our rendered datasets, we do not run any physics simulation so that we have complete control over the layout. We also render multiple query images for each object to be used in our instance detection task. To render a query image, we place an object into an empty background, rotate it along the z-axis and render an image every 45 degrees. Fig.~\ref{fig:images} shows some example target and query images.

\subsection{Pretraining with Instance Detection}
Given a query image depicting an object, our instance detection task is to detect the exact same instances of the object in the target image. 

\begin{figure*}[t]
    \centering
    \resizebox{0.8\textwidth}{!}{\includegraphics{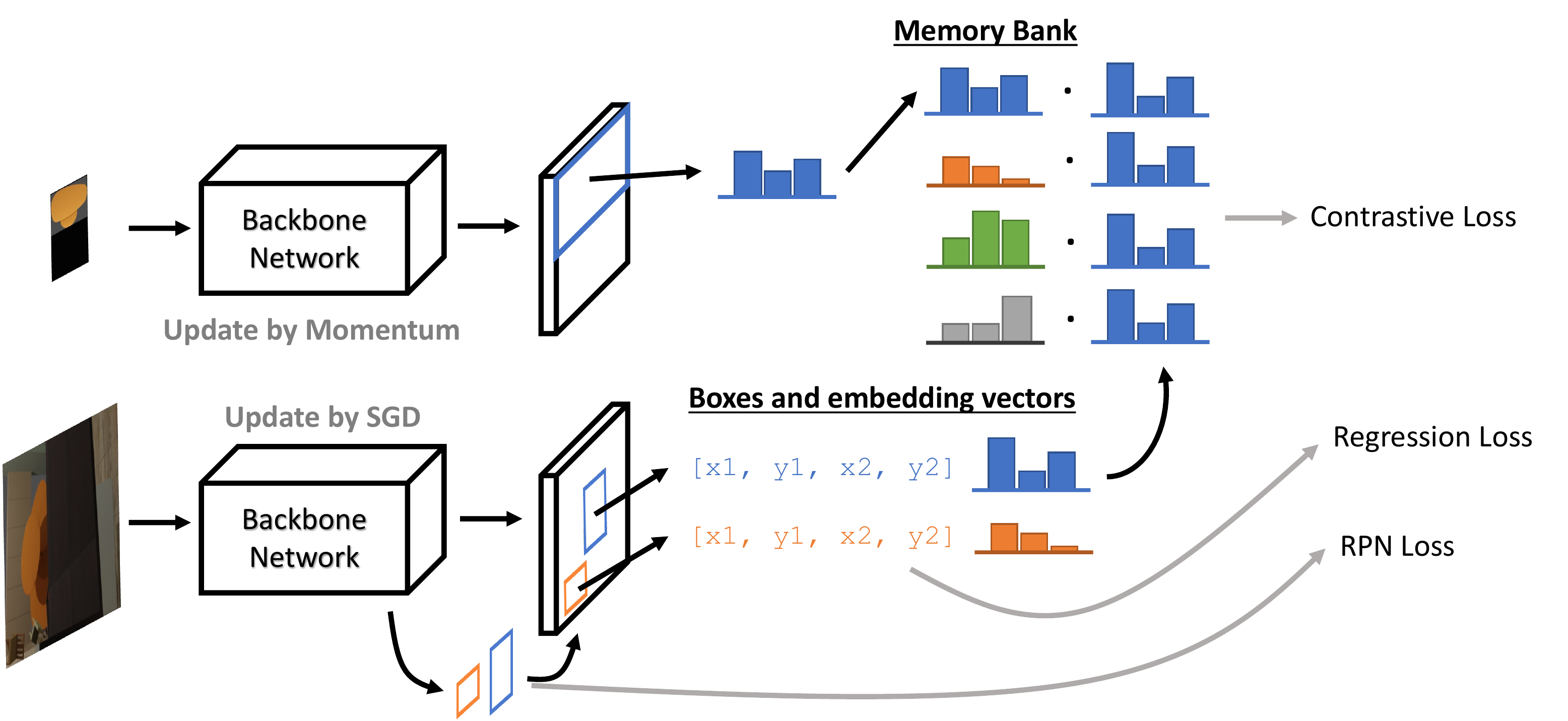}}
    \caption{Our network architecture consists of two detectors. The first detector is applied to a query image to extract an embedding vector for an object, which will be stored in a memory bank, and the second detector is applied to a target image to predict boxes and an embedding vector for each box. We calculate the dot products between an embedding vector of a region to all of the embedding vectors in the memory bank and apply contrastive loss, in addition to the standard regression loss and RPN loss. The first detector is updated by momentum while the second one is updated by SGD.}
    \label{fig:network}
    \vspace{-3mm}
\end{figure*}

Given an existing object detector, we propose a ``wrapper'' architecture to pretrain an object detector on our instance detection task as shown in Fig.~\ref{fig:network}. Our wrapper architecture consists of two identical object detectors, one for the query image and one for the target image. The first detector predicts an embedding vector for the query image, while the second detector detects objects in the target image and predicts an embedding vector for each predicted box. The embedding vectors should be similar if they depict the same object. We use Faster R-CNN and Mask R-CNN as our choice of detectors but other detectors such as one-stage detectors can be used in this wrapper architecture. Below, we use Faster R-CNN as an example to describe the details.

The first Faster R-CNN extracts an embedding vector $q_j$ from a query image depicting an object $j$ and does not predict any box. The query image is padded so that it can be fed into our network. We then use an RoIAlign layer to extract features for the object to not include features from the padded area. We also replace the classification head in this Faster R-CNN with a fully-connected layer with 256 channels which predicts an embedding vector from the features. Neither the RPN nor the bounding box regressor in this Faster R-CNN is used. Inspired by MoCo~\cite{DBLP:conf/cvpr/He0WXG20}, we have a memory bank that stores an embedding vector for every object.

There is a detail worth mentioning for updating the memory bank. We use multiple GPUs to train our detectors and each GPU has its own data sampling process. Each process chooses $n$ query images for its batch of target images independently. If there are more than $n$ unique objects in the target images, it randomly chooses $n$ of them and picks one random query image for each chosen object. Otherwise it samples from objects that are not in the target images. As a result each GPU uses different query images which creates inconsistency between memory banks of different GPUs. So before updating its memory bank, each GPU gathers the embedding vectors from other GPUs.

The second Faster R-CNN detects objects in target image. Similar to a conventional Faster R-CNN, it first predicts a set of region of interests (RoIs) and then uses RoIAlign layer to extract features for each RoI. Conventionally, the features are then used for predicting bounding box offsets and classes. Our wrapper architecture still predicts bounding box offsets but it predicts an embedding vector $k_i$ instead of a class for region $i$. We replace the classification layer with a fully-connected layer with 256 channels and the class-specific bounding box regressor with a class-agnostic one.

We then measure the similarities between a region and all objects by calculating dot products between an embedding vector of a region and all embedding vectors in the memory bank. We apply a contrastive loss function:
\begin{equation}
    \mathfrak{L}_{\text{con}} = -\sum_{i=1}^{M} \log \frac{\exp{(k_{i} \cdot q_{c_i} / \tau)}}{\sum_{j=1}^{N} \exp{(k_{i} \cdot q_{j} / \tau)}}
\end{equation}
where $M$ is the number of regions, $N$ is the number of objects, $c_i$ is the object for region $i$ and $\tau$ is a temperature hyper-parameter to train the detectors to predict similar embedding vectors for the same object. We follow MoCov2~\cite{DBLP:journals/corr/abs-2003-04297} to set $\tau$ to be 0.2 for all of our experiments. This loss is only applied to the foreground regions.

We use SGD to optimize the full training loss:
\begin{equation}
    \mathfrak{L} = \mathfrak{L}_{\text{con}} + \mathfrak{L}_{\text{reg}} + \mathfrak{L}_{\text{rpn}}
\end{equation}
where $\mathfrak{L}_{\text{reg}}$ is the bounding box regression loss and $\mathfrak{L}_{\text{rpn}}$ is the loss for RPN. And we use momentum~\cite{DBLP:conf/cvpr/He0WXG20,DBLP:journals/corr/abs-2003-04297} to update the parameters in Faster R-CNN for query images with a momentum coefficient of $0.999$ and the gradients to update the parameters in Faster R-CNN for target images.

\begin{table*}[ht]
\centering
\caption{We render multiple datasets of one million images each to demonstrate how pretraining data affects the performance of a detector. We also include SceneNet RGB-D for comparison. Fig.~\ref{fig:viewpoints} shows the viewpoint distributions of \texttt{Random Placement}, \texttt{Occlusion} and \texttt{Rotation}.\looseness=-1}
\resizebox{0.8\textwidth}{!}{
\begin{tabular}{l|ccc|cc|c}
Dataset            & \thead{Obj \\Count} & Occlu. & \thead{Scale Dist. \\(s / m / l)} & \thead{Rotation \\Axes} & Scene      & \thead{COCO \\AP} \\ \hline
\texttt{SceneNet RGB-D}~\cite{DBLP:conf/iccv/McCormacHLD17}       & 5.41      & -         & 45\% / 40\% / 15\%      & -             & SceneNet   & 36.6\%  \\ \hline
\texttt{Random Placement}     & 8.73      & 19\%      & 18\% / 52\% / 30\%      & Z axis        & White Cube & 37.2\%  \\
\texttt{Occlusion}            & 7.73      & 32\%      & 23\% / 48\% / 29\%      & Z axis        & White Cube & 37.2\%  \\
\texttt{Scale Distribution}   & 8.47      & 33\%      & 32\% / 37\% / 31\%      & Z axis        & White Cube & 37.5\%  \\
\texttt{Rotation}             & 8.10      & 33\%      & 35\% / 36\% / 29\%      & All axes      & White Cube & 37.7\%  \\
\texttt{SceneNet Background}  & 8.72      & 37\%      & 38\% / 34\% / 28\%      & All axes      & SceneNet   & 38.8\%  \\
\texttt{More Objects}         & 13.72     & 38\%      & 33\% / 39\% / 28\%      & All axes      & SceneNet   & 39.0\%  \\
\end{tabular}
}
\label{tab:datasets}
\vspace{-3mm}
\end{table*}

\begin{figure*}
\centering
\begin{subfigure}{0.30\textwidth}
\centering
\includegraphics[width=\linewidth]{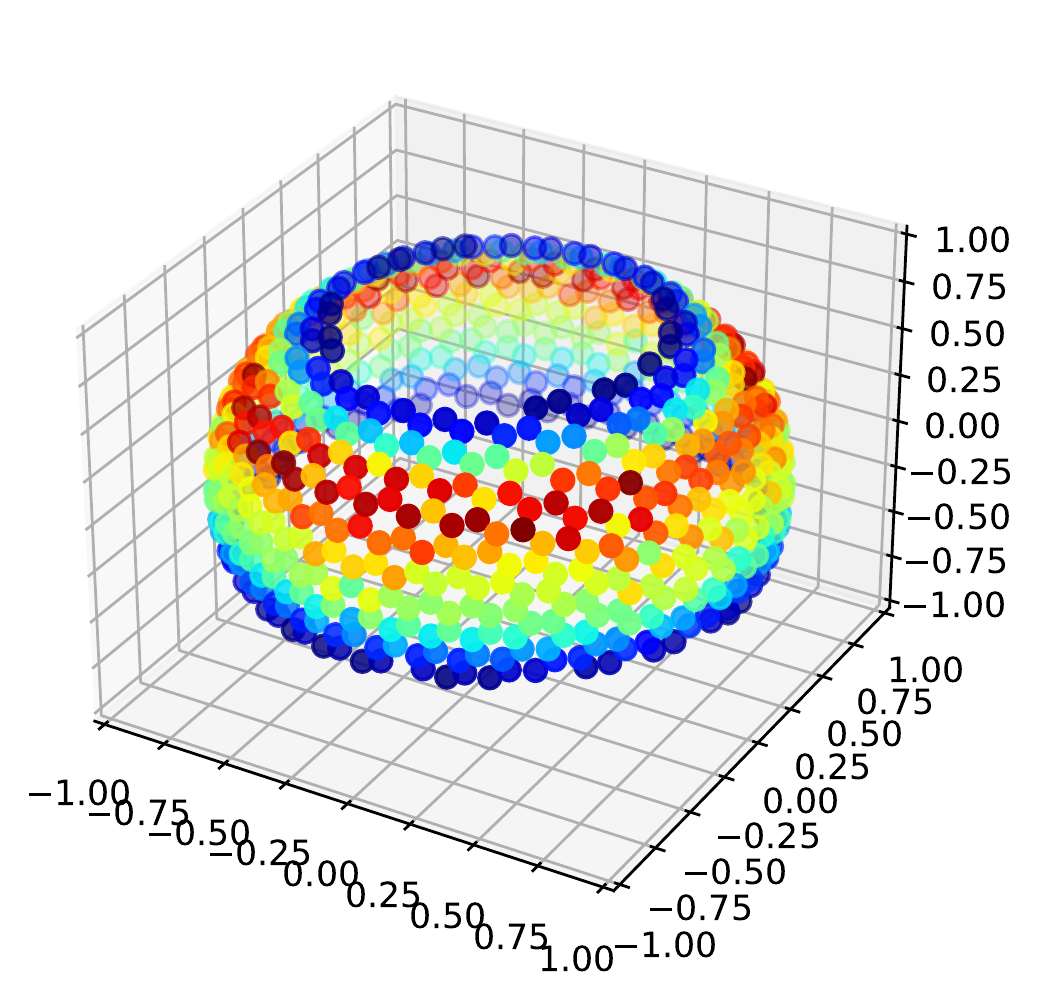}
\caption{\texttt{Random Placement}}
\end{subfigure}
\hfill
\begin{subfigure}{0.30\textwidth}
\centering
\includegraphics[width=\linewidth]{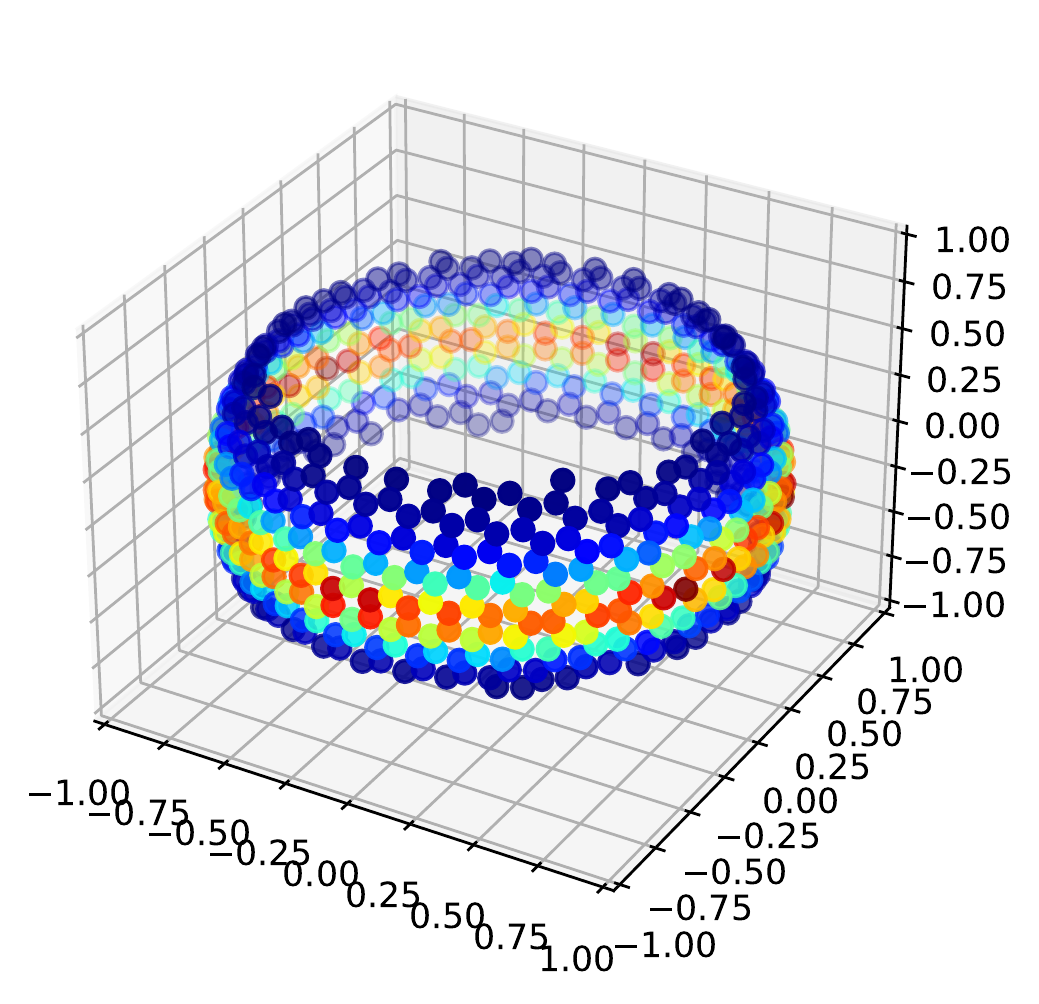}
\caption{\texttt{Occlusion}}
\end{subfigure}
\hfill
\begin{subfigure}[]{0.30\textwidth}
\centering
\includegraphics[width=\linewidth]{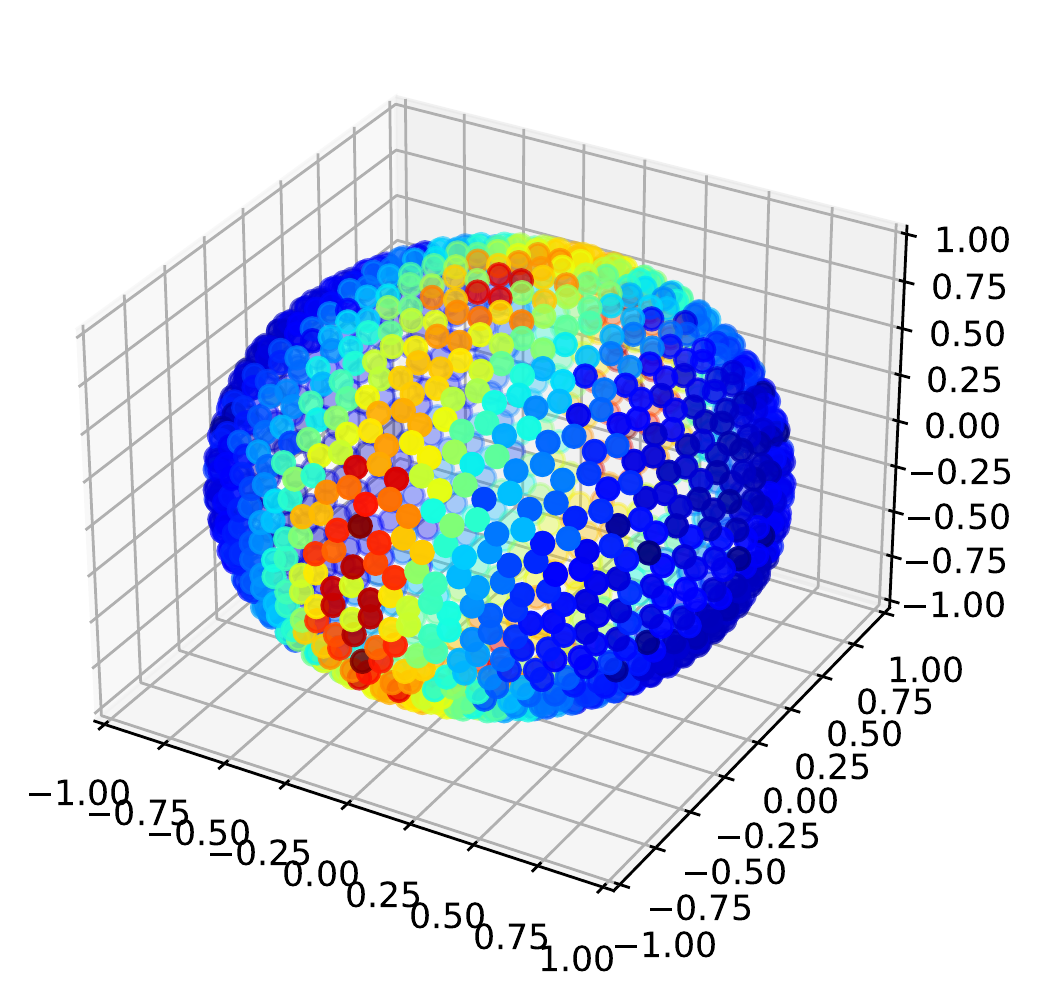}
\caption{\texttt{Rotation}}
\end{subfigure}
\caption{Viewpoint distribution changes with both the camera and object poses. In \texttt{Random Placement} and \texttt{Occlusion}, we only rotate along the z-axis so the top and bottom viewpoints are not well covered. In \texttt{Occlusion}, we restrict the camera poses so that it is easier to create occlusions between objects in the final images, which reduces the variations in viewpoints. In \texttt{Rotation}, we still restrict the camera poses but we rotate the objects along more axes to cover more viewpoints.}
\label{fig:viewpoints}
\vspace{-3mm}
\end{figure*}

\section{Experiments}
\label{sec:experiments}
\noindent \textbf{Implementation Details}~~We use 3D models from ShapeNet~\cite{DBLP:journals/corr/ChangFGHHLSSSSX15}, which can be used for non-commercial research, and indoor scenes from SceneNet~\cite{DBLP:journals/corr/HandaPBSC15a}, which is released under creative commons license, to construct our datasets. In our ablation studies, we use a subset of ShapeNet models that are used by SceneNet RGB-D~\cite{DBLP:conf/iccv/McCormacHLD17} and render images at $320 \times 240$. In experiments where we compare with existing approaches, we use all ShapeNet models and render images at $640 \times 480$.

We use Blender 2.92~\footnote{\url{https://www.blender.org}}, an open source 3D computer graphics software, and BlenderProc~\cite{DBLP:journals/corr/abs-1911-01911}, a Blender library, to render images and generate bounding box and mask annotations. We use the Cycles engine from Blender to render all images on GPUs. And we render one million images for each of the dataset in our experiments.

We implement our approach in Detectron2~\cite{wu2019detectron2} and use the default hyper-parameters, except the training schedule and input resolution, to pretrain our detectors. After pretraining, we initialize a downstream detector with weights in the detector for the target image excluding the classifier and regressor. To provide a fair comparison with other works, we fine-tune the detectors with the same schedule in each downstream setting, which will be described in detail in later sections.\looseness=-1

In our ablation experiments, during pretraining, we reduce the input resolution to half of the default input resolution. Our wrapper architecture is pretrained on 4 RTX3090 GPUs for 450k iterations with a batch size of 128 query images and a learning rate of 0.02. Each GPU samples 128 query images with a resolution of $112 \times 112$. When we compare our results with other approaches, we use the default input resolution in Detectron2 and increase the resolution of query images to $224 \times 224$. We optimize the learning rate schedule on COCO \texttt{val2017} under the 1x fine-tuning schedule. The wrapper architecture is pretrained on 8 A6000 GPUs for 750k iterations with an initial learning rate of 0.1 and a cosine annealing schedule~\cite{DBLP:conf/iclr/LoshchilovH17}. The target network uses batch normalization while the query network uses exponential moving average normalization~\cite{DBLP:conf/cvpr/CaiRMFTS21}, a variant of batch normalization designed for self-supervised learning. \looseness=-1

\begin{table*}[ht]
\centering
\caption{We compare our pretraining approach with existing pretraining approaches by fine-tuning a Mask R-CNN (\texttt{2fc}) with FPN and R-50 on COCO \texttt{train2017} under the standard 1x and 2x schedule, and evaluating it on COCO \texttt{testdev}. For each existing approach, we fine-tune a Mask R-CNN using the provided pretrained weight to get the test AP. We also include both the validation APs reported by each approach and validation APs reproduced by ourselves.}
\resizebox{0.9\textwidth}{!}{\begin{tabular}{l|cccccc|cccccc}
\multirow{2}{*}{}             & \multicolumn{6}{c|}{\textbf{1x Schedule}}                                                                                                                                                      & \multicolumn{6}{c}{\textbf{2x Schedule}}                                                                                                                                                      \\
                              & \multicolumn{3}{c}{$\text{AP}^{\text{bb}}$}                                                   & \multicolumn{3}{c|}{$\text{AP}^{\text{mk}}$}                                                   & \multicolumn{3}{c}{$\text{AP}^{\text{bb}}$}                                                   & \multicolumn{3}{c}{$\text{AP}^{\text{mk}}$}                                                   \\ \cline{2-13} 
                              & \multicolumn{1}{l}{\makecell{reported \\ \texttt{val}}} & \multicolumn{1}{l}{\makecell{reprod. \\ \texttt{val}}} & \multicolumn{1}{l}{\texttt{test}} & \multicolumn{1}{l}{\makecell{reported \\ \texttt{val}}} & \multicolumn{1}{l}{\makecell{reprod. \\ \texttt{val}}} & \multicolumn{1}{l|}{\texttt{test}} & \multicolumn{1}{l}{\makecell{reported \\ \texttt{val}}} & \multicolumn{1}{l}{\makecell{reprod. \\ \texttt{val}}} & \multicolumn{1}{l}{\texttt{test}} & \multicolumn{1}{l}{\makecell{reported \\ \texttt{val}}} & \multicolumn{1}{l}{\makecell{reprod. \\ \texttt{val}}} & \multicolumn{1}{l}{\texttt{test}} \\ \hline
Image Cls.                    & 38.9                             & -                               & -                        & 35.4                             & -                               & -                         & 40.6                             & -                               & -                        & 36.8                             & -                               & -                        \\
MoCov2~\cite{DBLP:journals/corr/abs-2003-04297}                        & 38.9                             & 39.6                            & 39.9                     & 35.4                             & 35.9                            & 36.2                      & 40.9                             & 41.6                            & 41.8                     & 37.0                             & 37.6                            & 37.9                     \\
SwAV~\cite{DBLP:conf/nips/CaronMMGBJ20}                          & -                                & 40.0                            & 40.5                     & -                                & 36.6                            & 37.2                      & -                                & 42.1                            & 42.5                     & -                                & 38.2                            & 38.7                     \\
DetCo~\cite{DBLP:conf/iccv/XieDWZXSL021}                         & 40.1                             & 40.0                            & 40.3                     & 36.4                             & 36.2                            & 36.6                      & -                                & 41.5                            & 42.0                     & -                                & 37.6                            & 38.1                     \\
$\text{ReSim-FPN}^{\text{T}}$~\cite{DBLP:conf/iccv/XiaoR0KD21} & 40.3                             & 40.4                            & 40.6                     & 36.4                             & 36.6                            & 36.8                      & 41.9                             & 41.8                            & 42.4                     & 37.9                             & 37.9                            & 38.4                     \\
MoCov3~\cite{DBLP:conf/iccv/ChenXH21}                        & -                                & 40.7                            & 41.1                     & -                                & 37.0                            & 37.5                      & -                                & 42.2                            & 42.6                     & -                                & 38.4                            & 38.6                     \\
DetCon\footnotemark~\cite{DBLP:conf/iccv/HenaffKAOVC21}                  & 42.7                             & 41.5                            & 41.8                     & 38.2                             & 37.6                            & 37.9                      & 43.4                             & 42.5                            & 42.8                     & 38.7                             & 38.4                            & 38.7                     \\
SCRL~\cite{DBLP:conf/cvpr/RohSKK21}                          & 41.3                             & 41.6                            & 42.0                     & 37.7                             & 37.5                            & 38.0                      & -                                & 42.8                            & 43.2                     & -                                & 38.7                            & 39.0                     \\
InsLoc~\cite{DBLP:conf/cvpr/YangWZL21}                          & 42.0                             & -                               & -                        & 37.6                             & -                               & -                         & 43.3                             & -                               & -                        & 38.8                             & -                               & -                        \\ \hline
Ours                          & -                                & 41.4                            & 41.5                     & -                                & 37.3                            & 37.5                      & -                                & 42.1                            & 42.8                     & -                                & 38.0                            & 38.6                    
\end{tabular}}
\label{tab:comparison}
\vspace{-3mm}
\end{table*}

\noindent \textbf{Synthetic Optimized Layout}~~We optimize the parameters of our rendering pipeline against a set of proxy metrics so that we maximize the effectiveness of the synthetic data. We render multiple sets of synthetic images with different proxy metrics and parameters. For each dataset, we pretrain a Faster R-CNN~\cite{DBLP:journals/pami/RenHG017} with a FPN~\cite{DBLP:conf/cvpr/LinDGHHB17} and ResNet-50~\cite{DBLP:conf/cvpr/HeZRS16}, fine-tune it on COCO~\texttt{train2017} under 1x schedule and evaluate it on COCO~\texttt{val2017}. Tab.~\ref{tab:datasets} shows the proxy metrics of each dataset and the corresponding validation performance. We also pretrain with SceneNet RGB-D~\cite{DBLP:conf/iccv/McCormacHLD17}, which is a large scale synthetic dataset and consists of 5 million images, in Tab.~\ref{tab:datasets} for comparison.

\noindent \texttt{Random Placement:} We start with randomly placing objects inside of a textureless cube. Each object is randomly scaled between $0.4$ and $2.0$ and rotated along the z-axis. The object is randomly placed in a location such that it does not collide with other objects and is on the floor. We randomly place a camera at a height between 0.1m and 5.0m, and point it toward a random point at a height between 0m and 2.0m in the central area of the background. This gives us mostly the eye-level and high-angle shots. The Faster R-CNN pretrained with this dataset achieves an AP of 37.2\% on COCO \texttt{val2017}.\looseness=-1 \\
\noindent \texttt{Occlusion:} Instead of random placement, the object is placed in front of or behind other objects with respect to the camera. The camera pose is adjusted to mostly eye-level shots to create occlusions between objects in the image. This increases the occlusion from $19\%$ to $32\%$ but reduces the viewpoint variation due to more constrained camera placement, as shown in Fig.~\ref{fig:viewpoints}. The AP stays at $37.2\%$. Later we show that increasing viewpoint variation improves the performance. \\
\noindent \texttt{Scale Distribution:} In the previous two configurations, an object is randomly scaled between $0.4$ and $2.0$ and the majority of the objects in the images are medium size. We divide the range into three intervals, $[0.1, 1.0]$, $[1.0, 2.0]$ and $[2.0, 3.0]$, and randomly select an interval with a probability of $0.7$, $0.1$ and $0.2$ respectively. This adds more small objects to the final dataset and improves the AP from $37.2\%$ to $37.5\%$. \\
\noindent \texttt{Rotation:} In this configuration, we not only rotate an object along the z axis but also the x and y axes. This includes more viewpoints of the objects in our dataset even if we are mostly using eye-level shots. Fig.~\ref{fig:viewpoints} shows how the viewpoint distributions change between \texttt{Scale Distribution} and \texttt{Rotation}. This dataset improves the AP from $37.5\%$ to $37.7\%$, which also explains why there is no improvement in \texttt{Occlusion}. \\
\noindent \texttt{SceneNet Background:} We use backgrounds from SceneNet and the AP improves from $37.7\%$ to $38.8\%$.\looseness=-1 \\
\noindent \texttt{More Objects:} We put more objects into the scene by allowing the objects to be floating. This increases the object count per image from 8.72 to 13.72 and the AP from $38.8\%$ to $39.0\%$.

The above experiments show that our proxy metrics are good indicators of validation performance. When we compare with existing pretraining approaches later, we build upon the \texttt{More Objects} configuration, using all 52k models from ShapeNet and rendering images at $640 \times 480$ instead of $320 \times 240$.\looseness=-1

\noindent \textbf{Instance Detection versus Alternative Pretraining Tasks}~~We evaluate the effectiveness of our label-free instance detection pretraining task by comparing it against alternative ways of pretraining including one that uses semantic labels. Using the \texttt{SceneNet Background} dataset, we compare our pretraining against two baseline methods to train the classifiers in the detector: (1) we treat each 3D model as an independent class and we have 21k classes in total; (2) we use the semantic labels in ShapeNet and group the 3D models into 148 categories. With the first approach, the training was unstable and diverged. With the second approach achieves an AP of $38.1\%$. In comparison, the network pretrained on our semantics-free instance detection task achieves an better AP of $38.8\%$.
\footnotetext[2]{DetCon originally uses the TPU implementation of Mask R-CNN instead of Detectron2 and different data augmentation during fine-tuning.}

\begin{table*}[ht]
\centering
\caption{We pretrain and fine-tune a Mask R-CNN (\texttt{4conv1fc}) with FPN and ResNet-50 by following the 400 epochs training schedule from the train-from-scratch baselines in Detectron2. We pretrain the Mask R-CNN on our pretraining task for the first half of the training schedule.}
\begin{tabular}{l|l|llllll}
pretrain                  & finetune & $\text{AP}^\text{bb}$   & $\text{AP}^\text{bb}_\text{50}$   & $\text{AP}^\text{bb}_\text{75}$   & $\text{AP}^\text{mk}$   & $\text{AP}^\text{mk}_\text{50}$   & $\text{AP}^\text{mk}_\text{75}$   \\ \hline
supervised & 3x schedule   & 41.9 & 61.7 & 45.8 & 37.8 & 59.2 & 40.5 \\
no & 9x schedule      & 43.6 & 63.6 & 47.6 & 39.3 & 60.9 & 42.3 \\
no & 100 epochs + LSJ~\cite{DBLP:conf/cvpr/GhiasiCSQLCLZ21} & 44.7 & 65.0 & 49.0 & 40.3 & 62.1 & 43.7 \\
no & 200 epochs + LSJ & 46.3 & 66.7 & 50.7 & 41.7 & 64.1 & 45.0 \\
no & 400 epochs + LSJ & 47.4 & 67.6 & 52.4 & 42.5 & 65.2 & 46.1 \\ \hline
Ours & 200 epochs + LSJ & 47.8 & 68.2 & 52.7 & 43.0 & 65.7 & 46.6
\end{tabular}
\label{tab:mask_rcnn_4conv1fc}
\end{table*}

\begin{table*}
\begin{minipage}{0.39\textwidth}
\caption{We fine-tune a Faster R-CNN with only 10\% of COCO \texttt{train2017}.}
\centering
\resizebox{0.8\textwidth}{!}{
\begin{tabular}{l|lll}
pretrain    & AP   & $\text{AP}^\text{50}$   & $\text{AP}^\text{75}$   \\ \hline
random      & 17.8 & 32.0 & 17.9 \\
ImageNet    & 22.6 & 38.4 & 23.5 \\ \hline
MoCov2      & 20.9 & 34.8 & 21.7 \\
SimCLRv2~\cite{DBLP:conf/nips/ChenKSNH20}    & 22.1 & 37.3 & 23.0 \\
SwAV        & 25.5 & 43.3 & 26.4 \\
BYOL~\cite{DBLP:conf/nips/GrillSATRBDPGAP20}        & 25.5 & 42.3 & 26.9 \\
SCRL        & 26.4 & 43.2 & 28.0 \\
SCRL (reprod.) & 26.7 & 43.3 & 28.3 \\ \hline
Ours        & 26.3 & 41.3 & 28.2
\end{tabular}}
\label{tab:low_data}
\end{minipage}
\hfill
\begin{minipage}{0.59\textwidth}
\caption{We evaluate our approach on the few-shot learning task.}
\centering
\resizebox{0.8\textwidth}{!}{
\begin{tabular}{l|cccccc}
\multirow{3}{*}{} & \multicolumn{6}{c}{10 Shot}                          \\
                  & \multicolumn{3}{c}{Base} & \multicolumn{3}{c}{Novel} \\
                  & AP     & AP50   & AP75   & AP      & AP50   & AP75   \\ \hline
ImageNet          & 32.9   & 52.2   & 36.3   & 8.8     & 17.4   & 7.9    \\
DetCo             & 34.4   & 54.2   & 37.8   & 9.5     & 18.0   & 8.9    \\
MoCov3            & 35.9   & 56.6   & 39.7   & 8.7     & 16.9   & 8.2    \\
$\text{ReSim-FPN}^\text{T}$         & 34.7   & 54.2   & 38.5   & 9.4     & 17.4   & 9.0    \\
SCRL              & 36.0   & 55.9   & 39.5   & 9.9     & 18.8   & 9.4    \\
SwAV              & 36.0   & 57.0   & 39.6   & 10.1    & 19.4   & 9.5    \\ \hline
Ours              & 35.9   & 55.0   & 39.7   & 9.4     & 17.3   & 8.9  
\end{tabular}}
\label{tab:few_shot}
\end{minipage}
\vspace{-3mm}
\end{table*}

\noindent \textbf{Comparisons with Existing Pretraining Approaches}~~To compare our approach with other pretraining approaches, we pretrain a Mask R-CNN (\texttt{2fc}) with an FPN and a ResNet-50 on our synthetic data. Following~\cite{DBLP:conf/cvpr/He0WXG20}, we then fine-tune it on COCO \texttt{train2017} under the standard 1x and 2x schedule and the same fine-tuning settings. In Tab.~\ref{tab:comparison}, in addition to the validation performance, we also include the test performance for a fair comparison as our learning schedule is optimized on the validation set under the 1x schedule. Since prior work only report validation performance, we fine-tune the network with the provided pretrained weights by ourselves to get the test performance. Our reproduced APs are similar to or better than the reported APs except for DetCon which originally uses the TPU implementation of Mask R-CNN and different data augmentations during fine-tuning. For the network pretrained on ImageNet, the TPU version achieves an AP of $39.6\%$ while the Detectron2 version achieves an AP of $38.9\%$. InsLoc modifies the architecture of Mask R-CNN by adding four convolution layers to the bounding box head, while other approaches use the architecture of Mask R-CNN in MoCo. We include the reported number (which is not directly comparable to other approaches) in Tab.~\ref{tab:comparison} for reference. Our approach achieves results competitive to the state-of-the-art pretraining approaches such as MoCov3, SCRL and DetCon. It is worth noting that our approach uses only 8 A6000 GPUs for pretraining while MoCov3 uses 16 V100 32G GPUs, SCRL uses 32 V100 GPUs and DetCon uses 128 TPU v3 workers.

\noindent \textbf{Pretraining versus Train-from-scratch}~~He et al.~\cite{DBLP:conf/iccv/HeGD19} show that a detector trained from scratch can be a strong baseline if it is trained long enough. Detectron2 provides a train-from-scratch baseline where it trains a Mask R-CNN (\texttt{4conv1fc}) with an FPN and a ResNet-50 and strong data augmentation~\cite{DBLP:conf/cvpr/GhiasiCSQLCLZ21} for 400 epochs on COCO from scratch. This baseline achieves an AP of 47.4\%, while the baseline pretrained on ImageNet only achieves an AP of 41.9\%. To verify that our pretraining still helps under a long fine-tuning schedule, we follow the 400-epochs training schedule where we use the first half for pretraining and fine-tune our network for 200 epochs. We use the same data augmentation and batch size. Tab.~\ref{tab:mask_rcnn_4conv1fc} shows that our approach outperforms the strong train-from-scratch baselines.

\noindent \textbf{Low Data Regime}~~Following SCRL~\cite{DBLP:conf/cvpr/RohSKK21}, we evaluate our approach in low data regime by fine-tuning the detector with only $10\%$ of COCO \texttt{train2017} data, as shown in Tab.~\ref{tab:low_data} which is adapted from SCRL. We also use the two-stage fine-tuning approach (TFA) from FsDet~\cite{DBLP:conf/icml/WangH0DY20} to evaluate our approach on the few-shot learning task, as shown in Tab.~\ref{tab:few_shot}. The 20 classes that are in both COCO and PASCAL VOC are used as novel classes while the 60 classes that are only in COCO are used as base classes. Each novel class has 10 training examples. TFA first fine-tunes the whole network on the base classes and then fine-tunes only the classifiers on the novel classes. Experiments show that our approach achieves results competitive to the state-of-the-art approaches.

\noindent \textbf{Limitations}~~Our ``SOLID'' approach mainly focuses on the spatial layout of the objects when generating synthetic images. But there are other aspects in the rendering pipeline that can potentially generate more effective pre-training data. Tab.~\ref{tab:datasets} shows that using backgrounds from SceneNet outperforms a textureless cube so it is possible that even more diversified backgrounds would be beneficial. Generating more photo realistic synthetic images may also reduce the domain gap between real and synthetic images.\looseness=-1

\section{Conclusion}
We have introduced SOLID, a new approach to pretraining an object detector with synthetic data. Our ``SOLID'' approach involves two main components: (1) synthetics images generated from a collection of unlablled 3D models with optimized scene arrangement; (2) pretraining an object detector on the ``instance detection'' task. Experiments on COCO show that synthetic data can be highly effective data for pretraining object detectors.\looseness=-1

\noindent \textbf{Acknowledgements}~~This work is partially supported by the Office of Naval Research under Grant N00014-20-1-2634.

{\small
\bibliographystyle{ieee_fullname}
\bibliography{egbib}
}

\end{document}